\documentclass[letterpaper, 10 pt, conference]{ieeeconf}  

\IEEEoverridecommandlockouts                              

\usepackage{graphicx}
\usepackage{caption}
\usepackage{booktabs}
\usepackage{subcaption}
\usepackage{url}

\title{\LARGE \bf
Autonomous Frontier-Based Exploration with VLM Guidance
}

\author{Aarush Aitha and Avideh Zakhor
\\ EECS Department, University of California
\\ {\tt\small \{aarush, avz\}@berkeley.edu}%
}

\begin{document}

\maketitle
\thispagestyle{empty}
\pagestyle{empty}

\begin{abstract}

Autonomous robotic exploration of unknown and hazardous environments, a long-standing challenge, can be significantly improved by leveraging the advanced reasoning of Vision-Language Models (VLMs). We introduce a novel exploration pipeline where a VLM performs high-level strategic decision-making, guiding a conventional low-level robotics control stack. At decision points, the robot generates a multimodal prompt with its current map and visual imagery of potential paths, or frontiers. The VLM analyzes this prompt to select the most promising frontier, replacing simple geometric heuristics with contextual spatial reasoning. This approach, validated in simulation across six indoor environments, improves map coverage by up to 24\% over existing methods. Our pipeline is lightweight, training-free, and easily transferable to any robot with standard sensors and an internet connection.

\end{abstract}

\section{INTRODUCTION}

For decades, the field of autonomous robotic exploration has relied on geometrically-driven algorithms such as frontier-based exploration \cite{yamauchi1997frontier} and Next-Best-View (NBV) planning \cite{bircher2016receding}. While foundational, these methods—and even sophisticated modern planners \cite{cao2021tare} \cite{zhu2021dsvp} share a critical weakness: their reliance on simple heuristics makes them prone to inefficient paths and strategically naive choices, as they lack a deeper, contextual understanding of the environment's structure. The recent and dramatic rise of Large Language Models (LLMs) and Vision-Language Models (VLMs) presents a transformative opportunity to bridge this gap \cite{chang2024survey}. These models possess an unprecedented ability to perform nuanced reasoning over multimodal data, including text and images. 
\\
This emergent capability for high-level spatial reasoning can be harnessed to overcome the limitations of purely geometric planners. We introduce a novel exploration pipeline that enhances a traditional frontier-based system with strategic guidance from a state-of-the-art VLM, Google's Gemini 2.5 Pro \cite{comanici2025gemini}. Our framework operates by delegating high-level decision-making to the VLM while relying on a robust, conventional robotics stack for low-level execution. 

In our proposed pipeline, when the robot encounters a decision point with multiple potential paths, it constructs a multimodal prompt. This prompt includes the current top-down occupancy map and visual imagery of the candidate frontiers. The VLM then analyzes this holistic data to select the single most promising frontier, not based on simple proximity, but on a strategic interpretation of the map's topology and visual cues. This approach effectively fuses the low-level navigational competence of the Robot Operating System (ROS) with the high-level strategic planning of a VLM - similar to how humans make explorational/navigational decisions. We demonstrate that integrating a VLM as a strategic commander leads to significantly more reliable and efficient exploration. Our proposed pipeline is lightweight, requires no model training, and is easily transferable to any robot with standard sensors and an internet connection. 


\section{RELATED WORK}

\subsection{Classical Geometric Methods}
\subsubsection{Frontier-Based Exploration}
A foundational strategy for autonomous exploration is frontier-based exploration, first proposed by Yamauchi \cite{yamauchi1997frontier}. This method iteratively navigates a robot to a frontier, defined as the boundary between mapped free space and unobserved territory, until no frontiers remain. The approach's simplicity and effectiveness have made it a benchmark. Later works such as the Wavefront Frontier Detector \cite{topiwala2018frontier} optimized the process of identifying frontier cells within an occupancy grid.

\subsubsection{Next-Best-View Planning}
Next-Best-View (NBV) planners sample a set of reachable viewpoints and select the one that maximizes an objective function, which is often the estimated volume of unknown space that would be revealed \cite{bircher2016receding}. NBV planners are typically more computationally intensive than basic frontier-based explorers as they must compute the objective function for every candidate viewpoint, which can be costly in 3D exploration.

\subsubsection{Hybrid Classical Approaches}
Recognizing the complementary strengths of these methods, hybrid hierarchical planners have been developed. Systems such as FUEL \cite{zhou2021fuel} and FAEP \cite{zhao2023autonomous} combine the global guidance of frontier-based methods with the local precision of NBV planning. Typically, a global planner identifies a promising frontier cluster, and a second-level planner refines this choice by selecting an optimal viewpoint within that cluster.

\subsection{Learning-Based and Language-Grounded Methods}
\subsubsection{Reinforcement Learning}
Reinforcement Learning (RL) has been applied to exploration by training a policy to select frontiers or actions that maximize information gain \cite{niroui2019deep, hu2020voronoi}. While RL-based agents can outperform classical methods, their primary drawback is the need for intensive data collection and training, often within simulation, which may not generalize perfectly to the real world. Our proposed approach, by contrast, is training-free.

\subsubsection{Language and Generative Models for Navigation}
More recently, the capabilities of large vision-language models (VLMs) have been leveraged to guide robotic navigation \cite{shah2023lm}. These approaches vary widely, using models to generate navigational subgoals \cite{bhorkar2024vlm}, determine when to stop exploring \cite{ren2024explore}, or compute navigability scores directly from images to influence motion commands \cite{zhang2025cliprover}.

\subsubsection{Language-Grounded Navigation for Object Finding (ObjectNav)}
Although the task of finding a specific object differs from pure exploration, its methods are relevant for using language models in spatial reasoning. A common paradigm is creating a semantically enriched spatial representation that a language model can reason over \cite{yokoyama2024vlfm, han2024hule, huang2022visual, chang2023goat, luo2024vlai}. These representations include language-grounded value maps that create a "heatmap" to guide the robot \cite{yokoyama2024vlfm, han2024hule}, language-aligned feature maps that can be "queried" with text \cite{huang2022visual}, or an explicit semantic memory of objects and locations \cite{chang2023goat, luo2024vlai}. Our work draws inspiration from this paradigm. However, we adapt it from the targeted task of finding a known object to the more open-ended challenge of general-purpose exploration. Rather than using a VLM to find a specific semantic target, we use its generalized spatial reasoning abilities to identify the most strategically promising direction to expand the map based on its topological structure.

\section{METHODOLOGY}

\subsection{Conceptual Overview of the Proposed Exploration Pipeline}

\begin{figure}[htbp]
    \centering
    \includegraphics[width=0.3\linewidth]{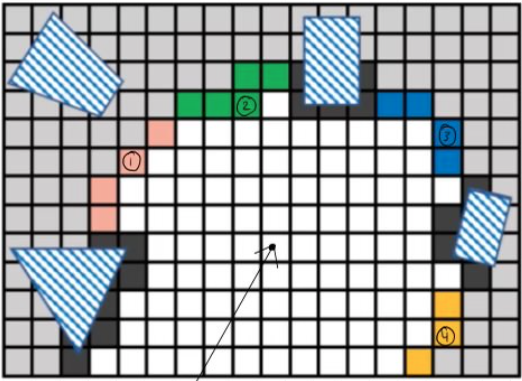}
    \captionof{figure}{An example of an occupancy map with four frontiers indicated in pink, green, blue, and yellow. White pixels are known free space, grey pixels are unknown space.}
    \label{fig:frontier_visualization}
\end{figure}

\begin{figure}[htbp]
    \centering
    \includegraphics[width=\linewidth]{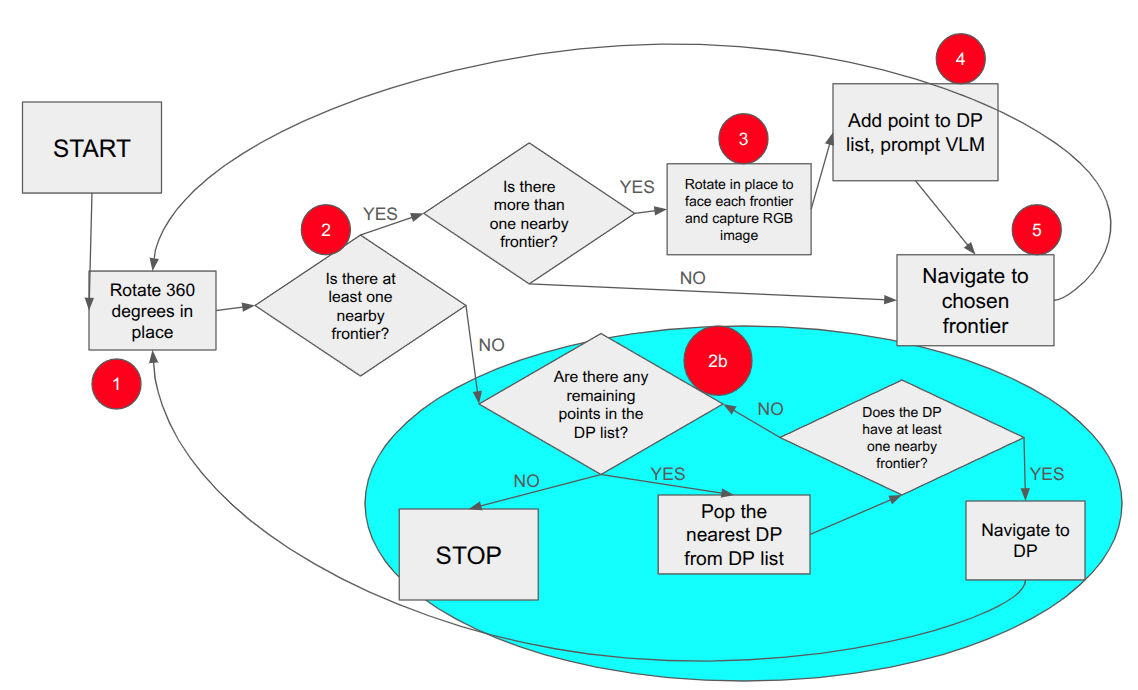}
    \captionof{figure}{A block diagram of our exploration pipeline.}
    \label{fig:block_diagram}
\end{figure}

The core claim of our work is that the semantic and spatial reasoning of a large VLM can enhance traditional frontier-based exploration, leading to more efficient and comprehensive mapping of unknown environments. To reiterate, a frontier is simply a boundary between free and unknown space on the occupancy map, as shown in Figure \ref{fig:frontier_visualization}. Our approach is an iterative decision-making loop that intelligently selects navigation goals. The summarized process is illustrated in Figure \ref{fig:block_diagram}.

The exploration cycle begins once the robot is initialized in an unknown space. First, the robot performs a 360-degree rotation to build an initial occupancy map of its immediate surroundings, as shown in Figure \ref{fig:block_diagram}, step 1. From this map, the system identifies frontiers, which represent all potential targets for expanding the map. Refer to Figure \ref{fig:frontier_visualization} for an example of an occupancy map with frontiers shown. To maintain focus and efficiency, these frontiers are filtered to consider only the most significant and proximate options, as shown in Figure \ref{fig:block_diagram} Step 2. For a demonstration of how this is implemented, see Figure \ref{fig:frontier_filtering}, to be discussed shortly in Section 3B.


At this stage, the robot's action depends on the number of viable frontiers.
\begin{itemize}
    \item If multiple frontiers are found, the robot has arrived at a ``decision point" (DP). Here, it pauses and engages the VLM to make an informed choice. Figures \ref{fig:frontier_filtering} and \ref{fig:frontier_blacklisting} both show examples of decision points, since in both situations the robot has multiple frontiers to choose from.
    \item If only one frontier is available, the robot proceeds autonomously by navigating towards it.
    \item If no frontiers are nearby, the robot has likely reached a dead-end. It then initiates a backtracking procedure by consulting a ``Decision Point List" (DP List) - a memory of previous locations with multiple exploration paths. See Figure \ref{fig:block_diagram}, step 2b for a visual explanation of this mechanism, and Section 3D for more details.
\end{itemize}


At a decision point, a multimodal prompt is constructed. This includes the current top-down 2D occupancy map, annotated with the robot's path and numerically labeled frontiers, along with images captured facing each frontier, as well as a textual prompt, an example of which is shown in Section 3C. This prompt is sent to Google's Gemini VLM, which is tasked with selecting the best frontier to explore next. The model returns its choice and reasoning, as shown in Figure \ref{fig:block_diagram}, step 4. An example of the VLM's response and spatial reasoning for choosing the frontier can be found in Figure \ref{fig:frontier_blacklisting}. The robot then computes a path and navigates toward the VLM-selected target as shown in Figure \ref{fig:block_diagram}, step 5, after which the entire cycle repeats. This continuous loop of mapping, identifying frontiers, and navigating systematically expands the known environment until there are no nearby frontiers left to explore and the decision point list is empty.

\subsection{Frontier Detection Strategy}

The foundation of our exploration algorithm is the robust detection of frontiers. We define a frontier cell as a known, free cell on the 2D occupancy grid that is adjacent to at least one unknown cell. To turn these individual cells into meaningful navigation goals, we first group contiguous frontier cells into clusters, or contours.

\begin{figure}[htbp]
    \centering
    \includegraphics[width=0.5\linewidth]{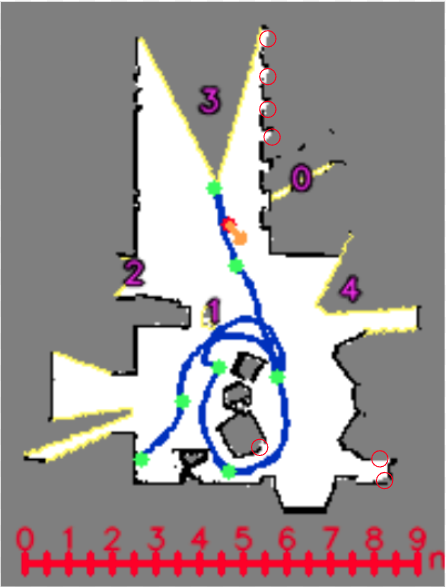}
    \captionof{figure}{Frontier filtering: The yellow lines represent frontiers that have passed the length check of 20 pixels, whereas the unmarked frontiers, circled in red, fail the length check. The labeled frontiers - (0), (1), (2), (3), and (4), shown in purple - are the five closest frontiers to the robot within a three meter radius. The blue line represents the robot's trajectory through the environment so far. The green dots represent points where the robot performed the 360 degree rotation as shown in Figure \ref{fig:block_diagram}, step 1. Finally, the robot's current position is marked by the red dot and its current heading is indicated by the orange arrow protruding from the red dot.}
    \label{fig:frontier_filtering}
\end{figure}

We achieve this using the \texttt{cv2.findContours} function from the OpenCV library, which processes a binary image of all identified frontier cells. This yields a list of distinct frontier contours. To ensure the robot pursues substantial areas of unknown space and is not distracted by mapping noise, we apply two filters. First, we discard small contours, e.g. those composed of fewer than 20 pixels, because the robot cannot easily traverse through these tight spaces. Second, we constrain the robot's focus to its immediate vicinity, e.g. by considering up to the five closest frontiers within a three-meter radius. This spatial constraint prevents long, inefficient traversals across the map and focuses the VLM's decision-making on relevant, nearby options. Both filtering operations are shown in Figure \ref{fig:frontier_filtering}.

\subsection{VLM-Guided Frontier Selection}

The central hypothesis of our work is that a VLM can introduce human-like strategic reasoning into the exploration process, moving beyond simple geometric heuristics. At each decision point, we employ Google's Gemini 2.5 Pro model \cite{comanici2025gemini} as our decision-making engine.

A multimodal prompt is generated, containing:
\begin{enumerate}
    \item Visual Context: A single image of the current 2D occupancy map, showing the robot's traversal history and the candidate frontiers clearly marked with numerical labels. Figure \ref{fig:frontier_filtering} shows an example of this.
    \item Textual Instruction: A task-oriented prompt: ``You are an autonomous exploration robot. Your goal is to fully explore/cover the entire environment as quickly and efficiently as possible. You are given the current top-down occupancy map and a series of images. Each image corresponds to a numbered 'frontier' on the map. Based on the images, the map, and previous steps, choose the single best frontier to navigate towards to maximize exploration efficiency while minimizing total distance traveled. Your answer must be only the number of the chosen frontier (e.g., '0', '1', etc.) with absolutely nothing else. Starting on the next line directly after this, please explain your reasoning as to how you chose the frontier."

\end{enumerate}

In our preliminary evaluations, we selected Gemini 2.5 Pro after comparing its qualitative performance against OpenAI's GPT-4o \cite{achiam2023gpt}. Gemini 2.5 Pro consistently demonstrated superior spatial and strategic reasoning. It is more adept at interpreting the abstract occupancy grid layouts and providing choices that align with intelligent exploration heuristics, such as prioritizing frontiers that appeared to lead into large, open spaces or those that would resolve major ambiguities in the map structure.
\\
A key technical advantage that informed our choice is Gemini 2.5 Pro's large context window and ability to maintain a continuous conversational session. We leverage this feature to create a persistent chat session for the duration of an exploration run. This provides the VLM with a memory of its previous decisions and their outcomes, allowing for more informed decisions. For instance, the model could be reminded of a previously chosen dead-end, potentially influencing it to make a more adventurous choice in a subsequent step. This capability elevates the VLM from a simple classifier to an exploration guider with contextual awareness, which is critical for complex, long-term autonomous tasks.

\subsection{Ensuring Complete and Robust Exploration}

To handle navigation failures and ensure the entire environment is mapped without getting trapped, our pipeline incorporates two critical mechanisms: a Decision Point List for systematic backtracking and a frontier blacklisting system for robustness.

\subsubsection{The Decision Point List for Systematic Backtracking}

To guarantee full map coverage and prevent the robot from becoming permanently stuck in areas with no immediate frontiers, we implement a backtracking mechanism using a Decision Point List. This list serves as a memory of locations with unexplored potential. A location is added to this list whenever the robot identifies more than one valid frontier in its vicinity. When the robot finds itself with no nearby frontiers to explore, it then sorts the list by Euclidean distance from its current position and attempts to navigate to the closest decision point. This nearest-first strategy minimizes backtracking travel time. If that point no longer has open frontiers upon arrival, it is removed from the list, and the robot proceeds to the next closest decision point. The entire exploration task is considered complete when this list becomes empty.

\subsubsection{Frontier Blacklisting for Navigation Robustness}

\begin{figure*}[htbp]
    \centering
    \includegraphics[width=0.9\textwidth]{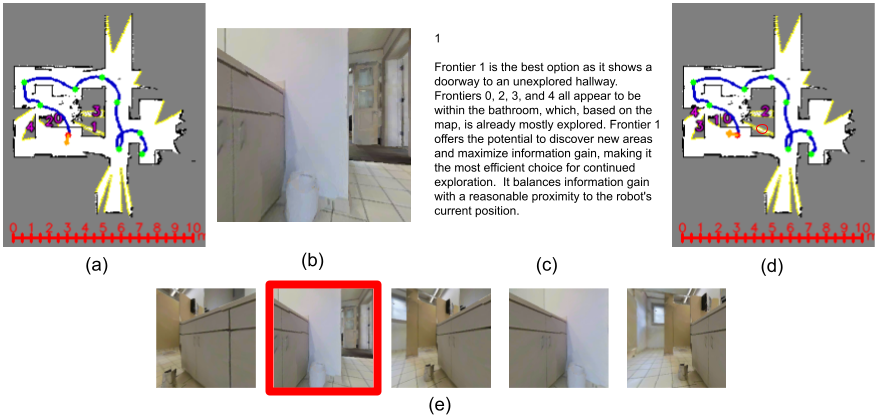}
    \caption{Frontier Blacklisting: (a) the occupancy map with multple frontiers, (b) the RGB image of the chosen frontier, (c) the VLM's reasoning for choosing frontier 1, (d) the subsequent occupancy map with frontier 1 having been blacklisted due to being unreachable by the local navigation policy, and (e) the images of frontiers 0, 1, 2, 3, and 4 from left to right, with the selected frontier bordered in red.}
    \label{fig:frontier_blacklisting}
\end{figure*}

Occasionally, the navigation planner may fail to find a valid path to a chosen frontier, perhaps because the goal is physically unreachable or inside an obstacle not yet fully mapped. To prevent the system from repeatedly attempting to navigate to a known-impossible goal, we implement a frontier blacklisting mechanism. As illustrated in Figure \ref{fig:frontier_blacklisting}, if the navigation stack fails to generate a path to the chosen frontier, the midpoint of that frontier is added to a blacklist. In all subsequent iterations, any frontier whose midpoint falls within a fixed radius of a blacklisted point is ignored. This ensures the robot does not waste time and resources on futile navigation attempts and encourages it to consider other options. This same mechanism is also used to blacklist a frontier after it has been successfully reached, preventing the algorithm from re-selecting a region that has already been explored.

\subsubsection{Local Navigation}

When navigating to a goal point, first a global path to the selected frontier is determined by \texttt{move\_base}, which is a ROS navigation framework integrating global and local planners, costmaps, and recovery behaviors to generate feasible paths \cite{movebase2020}. The final execution of the trajectory is handled by a local planner. For this critical task, we utilize the Timed Elastic Band (TEB) local planner \cite{rosmann2015timed}, a ROS package designed for real-time path optimization and navigation in constrained spaces.

\subsection{Implementation Framework}

Our system is realized through a combination of a high-fidelity simulator and a suite of standard robotics software, enabling rapid development and repeatable testing.

\subsubsection{Simulation with Habitat and Matterport3D}

To accelerate the development cycle and enable rigorous, repeatable testing, we adopted a simulation methodology. Our simulation framework is built on Habitat \cite{savva2019habitat}, utilizing the Matterport3D dataset \cite{chang2017matterport3d} for its large-scale, photorealistic environments. Crucially, we leveraged the ROS-x-Habitat package \cite{chen2022ros} to bridge the gap between the simulation platform and our robotics software stack. The ROS-x-Habitat bridge allows an agent within the Habitat environment to publish its sensor data, such as RGB-D camera data and odometry, directly to ROS topics. This enables our entire ROS-based pipeline, including Real-Time Appearance Based Mapping (RTAB-Map) \cite{labbe24rtab} and our exploration planner, to operate on the simulated data as if it were deployed on a physical robot, with no modifications to the core code.


\subsubsection{SLAM and Navigation with ROS and RTAB-Map}

Our software architecture is built on ROS Noetic running on Ubuntu 20.04. For SLAM, we use RTAB-Map \cite{labbe24rtab}.


In order to generate the occupancy map, we take advantage of the \texttt{depthimage\_to\_laserscan} ROS package. This package essentially projects a 40 pixel wide strip from the middle of the current depth image onto the top-down occupancy map as a laser scan, which is then passed to RTAB-Map to stitch together the scans to create the overall occupancy map.

\section{RESULTS}

\subsection{Baseline Methodologies}
To contextualize the performance of our VLM-driven approach, we selected four diverse and representative baseline methods from the existing literature: (a) Greedy Frontier-Based Exploration (Explore Lite) \cite{Hörner2016}; (b) OpenCV + NBV (Next-Best-View) \cite{umari2017autonomous}, a classic Next-Best-View planner using OpenCV to detect frontiers; (c) Technologies for Autonomous Robot Exploration (TARE) \cite{cao2021tare}, a sophisticated hierarchical framework designed for efficient 3D exploration; and (d) Dual-Stage Viewpoint Planner (DSVP) \cite{zhu2021dsvp}, a planner that explicitly separates the exploration task into two stages: a local exploration stage that uses a dynamically expanding RRT to extend the map boundary, and a global relocation stage that uses a graph to transit the robot between different sub-areas.

Given that TARE and DSVP are designed to run within Carnegie Mellon University’s
Autonomous Exploration Development Environment (AEDE) and as such, cannot be
easily simulated in Habitat, we need to devise a proper evaluation methodology to compare their performance with our method and the other two baseline methods that natively run on Habitat. To do so, we opted to capture the trajectories of TARE and DSVP by running them on AEDE, and then using the trajectories to evaluate them within the same ROS-x-Habitat ecosystem that was used for the other methods. 
\\
Concretely, for the specific purpose of evaluating exploration coverage of all methods in ROS-x-Habitat, we employ a privileged method within the Habitat simulator. This method updates a 2D top-down map by revealing a 360°, 5-meter radius around the agent’s ground-truth position at each step. This evaluation mechanism serves as an idealized measure of the explored area, simulating a perfect, circular ``sight” for metric calculation. Examples of such evaluations are shown in Figures \ref{fig:final_pics_env_1} and \ref{fig:final_pics_env_2}, where the dark gray areas denote unexplored areas. For completeness, we also evaluate with an FoV of 90° \cite{aitha2025frontier} and reach similar conclusions. To evaluate the TARE and DSVP methods, we considered a 360° field of view with evaluation ranges of 5 m and 20 m. Increasing the evaluation range to 20 m resulted in a significant reduction in rendering speed across all simulations and also produced negligible differences in the outcome. Since the environments are bounded to at most 25 × 25 m, it is reasonable to conclude that enlarging the evaluation range beyond 20 m - e.g. up to 100 m, which is the range used by TARE and DSVP in AEDE - would not yield substantially different results. Accordingly, we adopt a 360° field of view with a 5 m evaluation range for all experiments.

\subsection{Comparative Analysis of Results}

\subsubsection{Summary of Performances}

Table \ref{tab:comparison_wrapped_360} compares the ``final" distance traveled and exploration percentage of our method against the four baseline methods discussed earlier. In these tables, we use the average performance across 3 runs for all environments for our method. Specifically, Table \ref{tab:comparison_wrapped_360} shows that our VLM-guided method is reliable, achieving over 90\% coverage in all six environments and surpassing 98\% in four of them. It reaches the highest final exploration percentage in all six environments, demonstrating a clear advantage in completeness.

\begin{table}[ht!]
\centering
\caption{Total Distance Traveled vs. Final Exploration Percentage}
\label{tab:comparison_wrapped_360}
\scriptsize
\setlength{\tabcolsep}{3pt} 

\begin{tabular}{@{}l*{6}{c}@{}} 
\toprule
& \multicolumn{2}{c}{\textbf{Env. 1}} & \multicolumn{2}{c}{\textbf{Env. 2}} & \multicolumn{2}{c}{\textbf{Env. 3}} \\
\cmidrule(lr){2-3} \cmidrule(lr){4-5} \cmidrule(lr){6-7}
\textbf{Method} & \textbf{Dist.} & \textbf{Expl.} & \textbf{Dist.} & \textbf{Expl.} & \textbf{Dist.} & \textbf{Expl.} \\
\midrule
Ours                  & 43.48 & 98.99         & 468.51 & 99.89         & 57.26 & 99.24         \\
Greedy \cite{Hörner2016}    & 34.32 & 82.45         & 517.40 & 90.16         & 24.51 & 95.93         \\
NBV \cite{umari2017autonomous} & 90.52 & 90.10         & 734.46 & 93.12         & 45.67 & 95.33         \\
TARE \cite{cao2021tare}       & 39.35 & 93.14         & 188.42 & 87.69         & 30.25 & 96.11         \\
DSVP \cite{zhu2021dsvp}       & 40.60 & 96.53         & 213.55 & 92.26         & 40.93 & 96.69         \\
\bottomrule
\end{tabular}

\vspace{1.5em}

\setlength{\tabcolsep}{3pt} 
\begin{tabular}{@{}l*{6}{c}@{}} 
\toprule
& \multicolumn{2}{c}{\textbf{Env. 4}} & \multicolumn{2}{c}{\textbf{Env. 5}} & \multicolumn{2}{c}{\textbf{Env. 6}} \\
\cmidrule(lr){2-3} \cmidrule(lr){4-5} \cmidrule(lr){6-7}
\textbf{Method} & \textbf{Dist.} & \textbf{Expl.} & \textbf{Dist.} & \textbf{Expl.} & \textbf{Dist.} & \textbf{Expl.} \\
\midrule
Ours                  & 161.73 & 99.89        & 259.62 & 99.33         & 133.30 & 94.11         \\
Greedy \cite{Hörner2016}    & 207.66 & 96.77        & 350.75 & 89.62         & 125.61 & 86.12         \\
NBV \cite{umari2017autonomous} & 206.49 & 91.05        & 391.55 & 96.42         & 87.01  & 69.14         \\
TARE \cite{cao2021tare}       & 123.48 & 83.87        & 80.78  & 68.64         & 125.50 & 88.83         \\
DSVP \cite{zhu2021dsvp}       & 105.65 & 83.98        & ------ & -----         & 84.25  & 79.75         \\
\bottomrule
\end{tabular}
\end{table}

\begin{figure}[htbp!]
    \centering
    \includegraphics[width=0.8\linewidth]{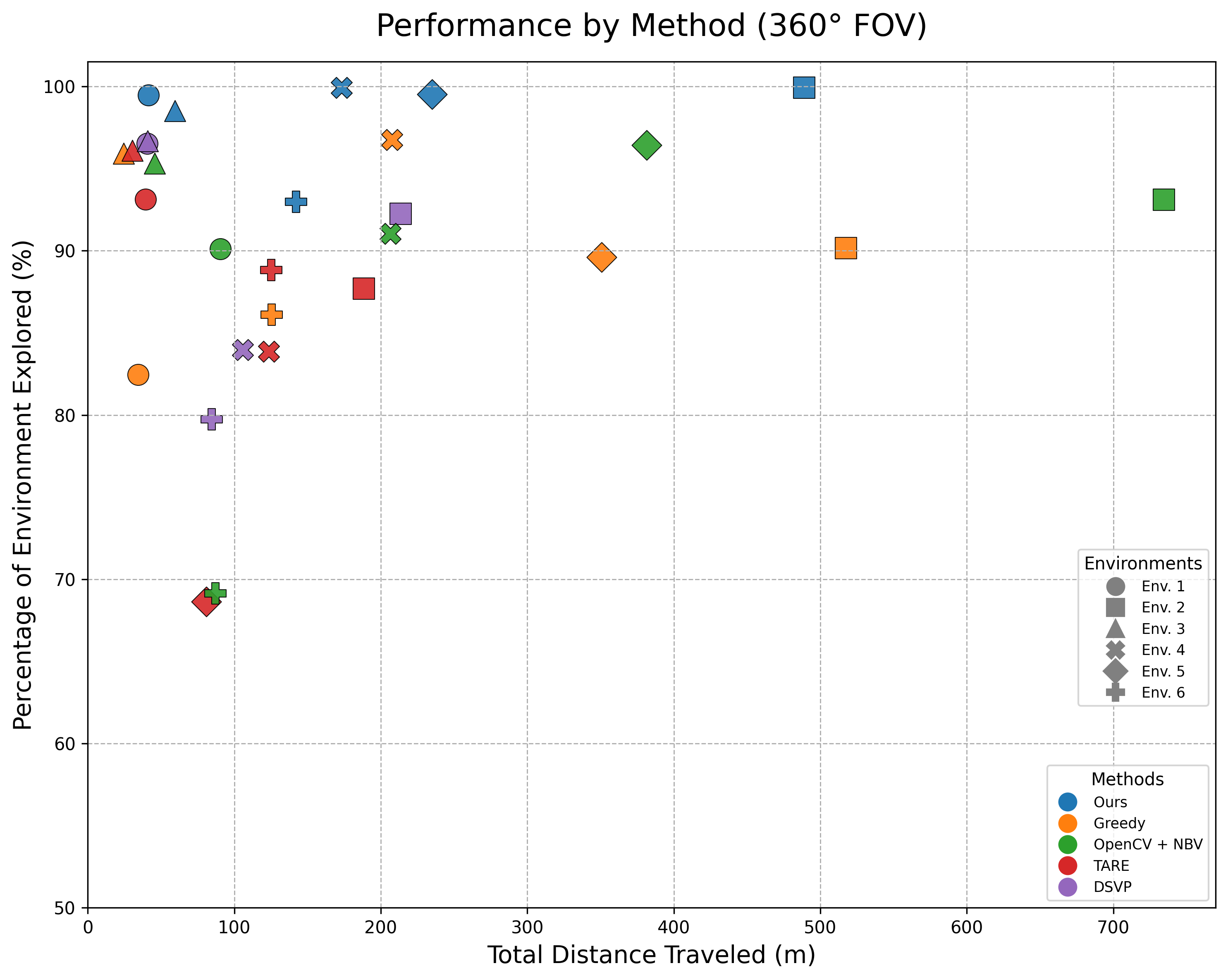}
    
    
    \caption{Final Exploration Percentage vs. Total Distance Traveled}
    \label{fig:symbol_plots}
\end{figure}

Figure \ref{fig:symbol_plots} visualizes the final performance data of Table \ref{tab:comparison_wrapped_360} as a scatter plot, comparing exploration percentage against the total distance traveled. In this plot, the ideal performance is located in the top-left corner, i.e. maximum exploration for minimum distance. For nearly every environment, our method is positioned favorably in the top-left portion of the graph, achieving the highest exploration percentage while relatively minimizing distance traveled. As shown later, while some methods such as TARE and DSVP occasionally travel shorter distances, this is a direct result of their premature termination, as they fail to explore the entire map.
\\

\subsubsection{Exploration Efficiency over Time}

\begin{figure}[htbp!]
    \centering
    \begin{subfigure}{0.49\linewidth}
        \centering
        \includegraphics[width=\linewidth]{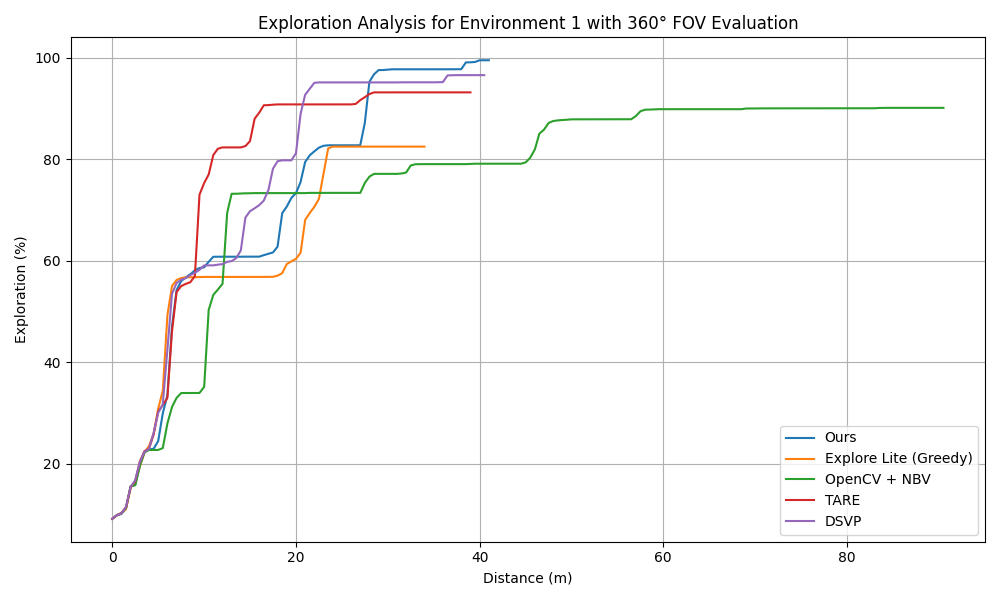}
        \caption{}
    \end{subfigure}
    \hfill 
    \begin{subfigure}{0.49\linewidth}
        \centering
        \includegraphics[width=\linewidth]{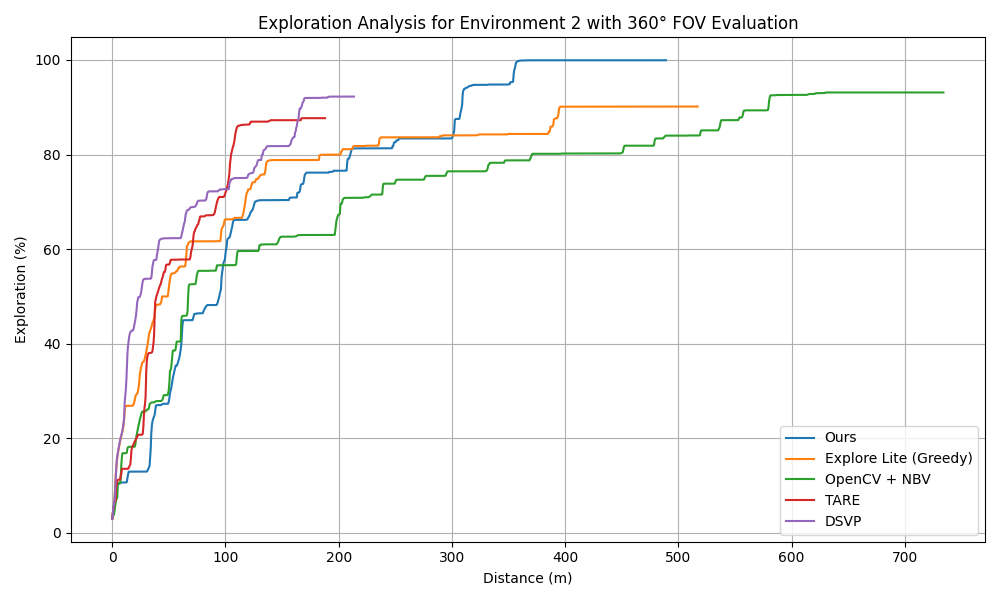}
        \caption{}
    \end{subfigure}

    \begin{subfigure}{0.49\linewidth}
        \centering
        \includegraphics[width=\linewidth]{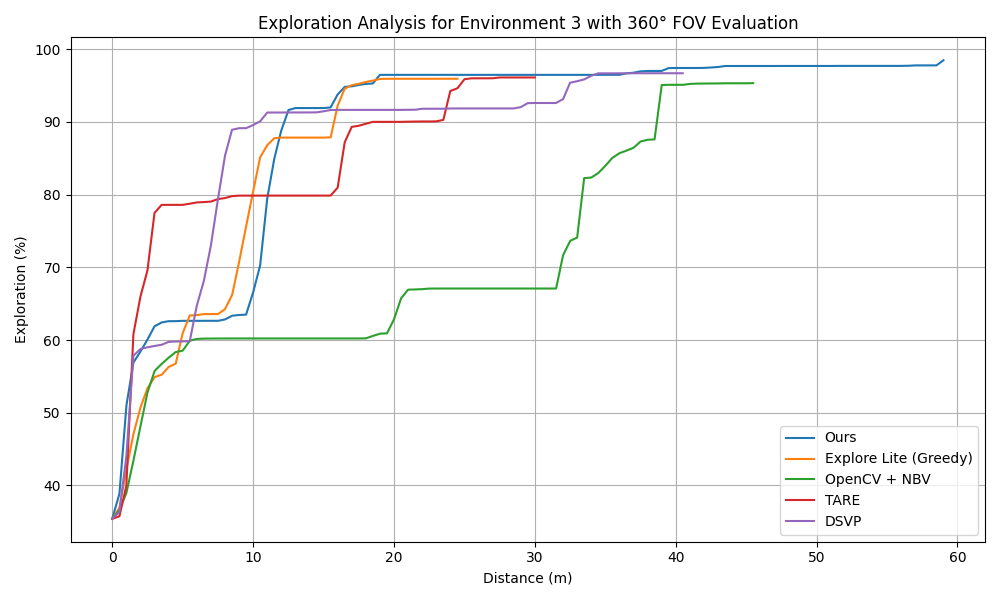}
        \caption{}
    \end{subfigure}
    \hfill
    \begin{subfigure}{0.49\linewidth}
        \centering
        \includegraphics[width=\linewidth]{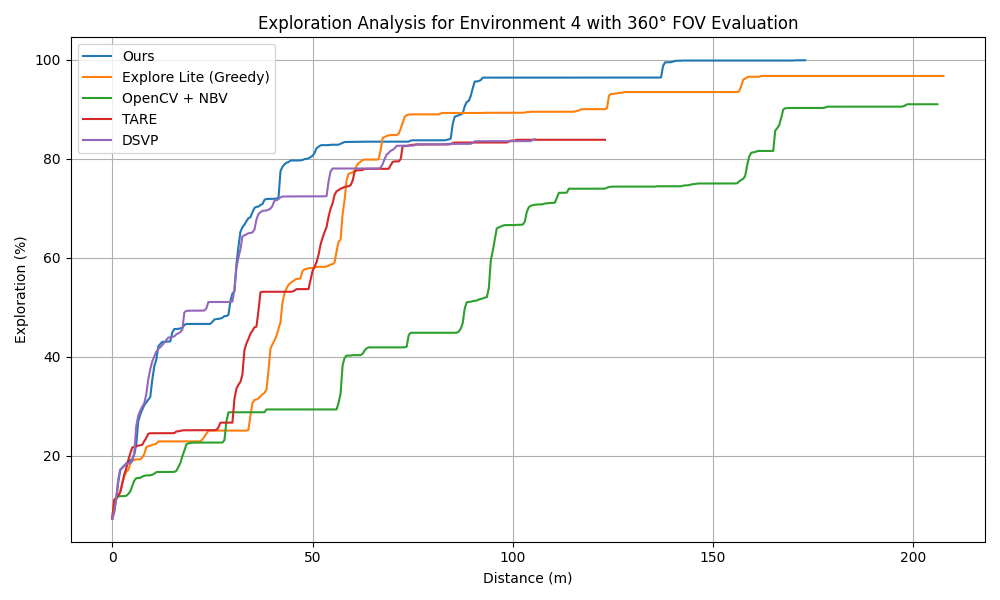}
        \caption{}
    \end{subfigure}

    \begin{subfigure}{0.49\linewidth}
        \centering
        \includegraphics[width=\linewidth]{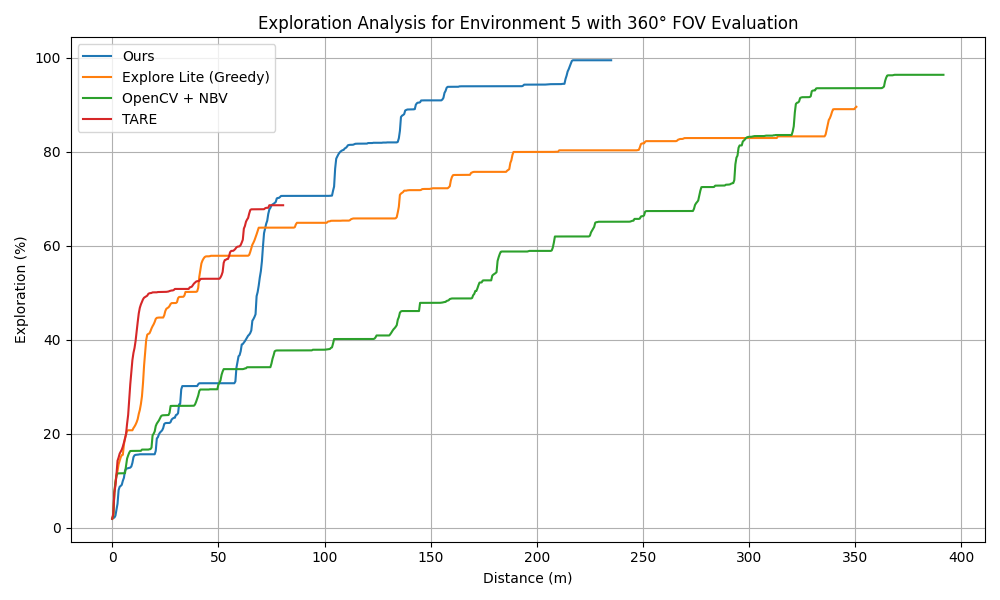}
        \caption{}
    \end{subfigure}
    \hfill
    \begin{subfigure}{0.49\linewidth}
        \centering
        \includegraphics[width=\linewidth]{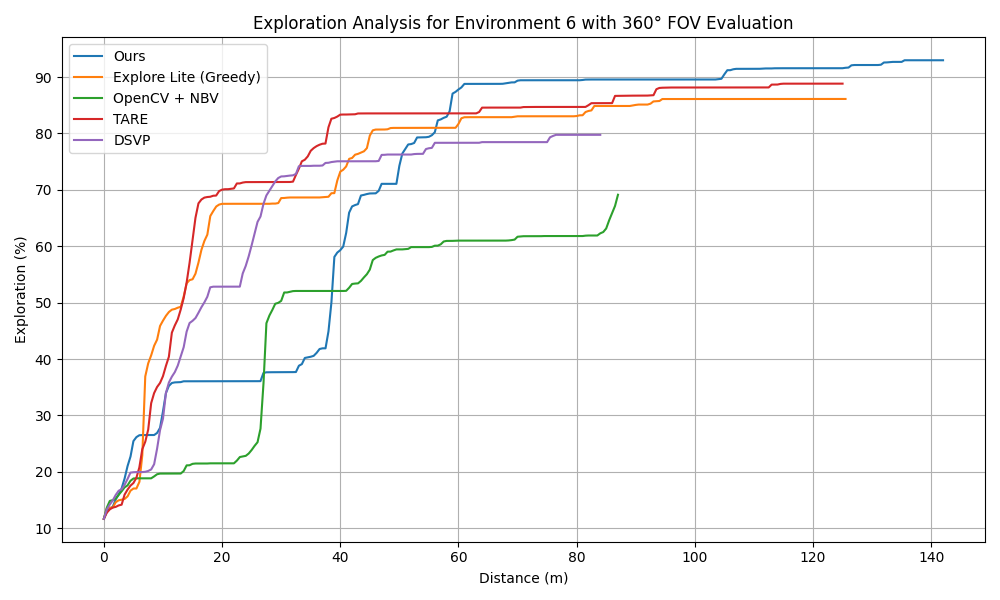}
        \caption{}
    \end{subfigure}
    
    \caption{Exploration analysis plots for environments (a) one, (b) two, (c) three, (d) four, (e) five, and (f) six.}
    \label{fig:exp_vs_dist_line_360}
\end{figure}

To analyze not only the final outcome but the efficiency of the exploration process itself, we turn to the progression plots in Figure \ref{fig:exp_vs_dist_line_360}, which show the exploration percentage versus distance traveled over the entire run, where a steeper curve indicates a more efficient exploration strategy - i.e., more of the map is discovered per meter traveled. In nearly every case, the curve representing our method shows a steeper initial slope and a more consistent climb. This demonstrates that the VLM's strategic choices lead to a more rapid discovery of new space from the outset, wasting less time on suboptimal paths. In contrast, the curves for the Greedy and NBV planners frequently flat-line, indicating long periods of travel with no new information gain. Environment 2, as shown in Figure \ref{fig:exp_vs_dist_line_360}(b), provides a compelling case study. Referencing the data from Table \ref{tab:comparison_wrapped_360}, our method achieved 99.46\% coverage after traveling 426.71 meters. The Greedy method, however, traveled nearly 100 meters farther - 517.40 m - and still only mapped 90.16\% of the area.
\\

\subsubsection{Qualitative Analysis of Exploration Trajectories}

\begin{figure}[htbp!]
    \begin{subfigure}{0.4\linewidth}
        \centering
        \includegraphics[width=\linewidth]{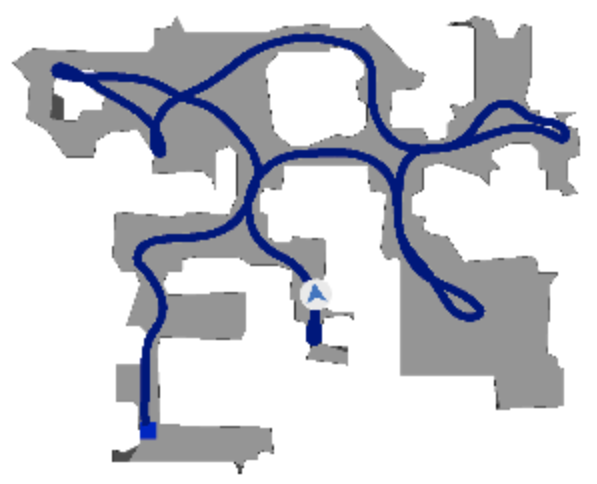}
        \caption{}
    \end{subfigure}
    \hfill
    \begin{subfigure}{0.4\linewidth}
        \centering
        \includegraphics[width=\linewidth]{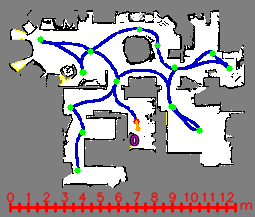}
        \caption{}
    \end{subfigure}
    \begin{subfigure}{0.4\linewidth}
        \centering
        \includegraphics[width=\linewidth]{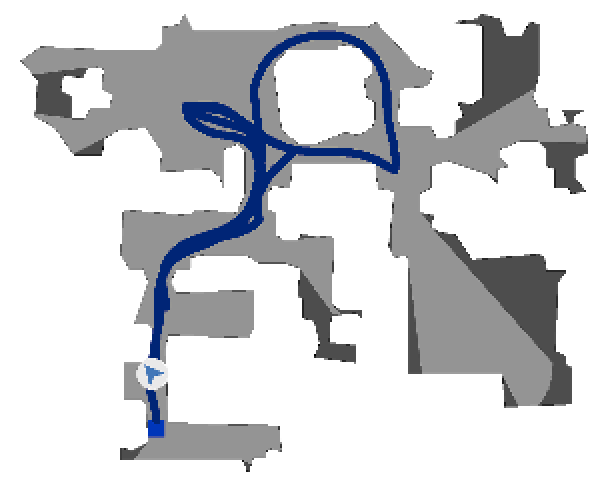}
        \caption{}
    \end{subfigure}
    \begin{subfigure}{0.4\linewidth}
        \centering
        \includegraphics[width=\linewidth]{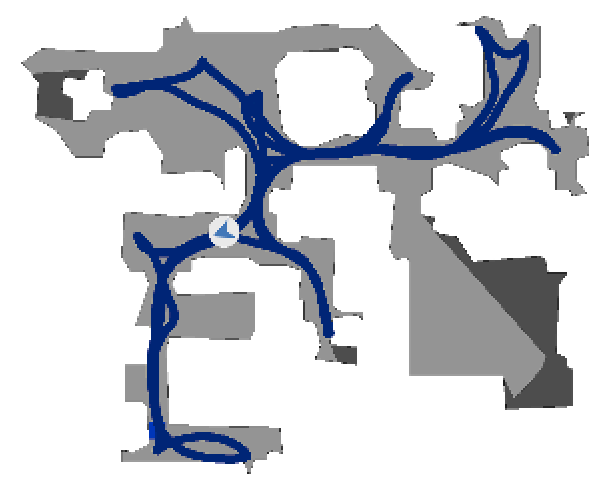}
        \caption{}
    \end{subfigure}
    \begin{subfigure}{0.4\linewidth}
        \centering
        \includegraphics[width=\linewidth]{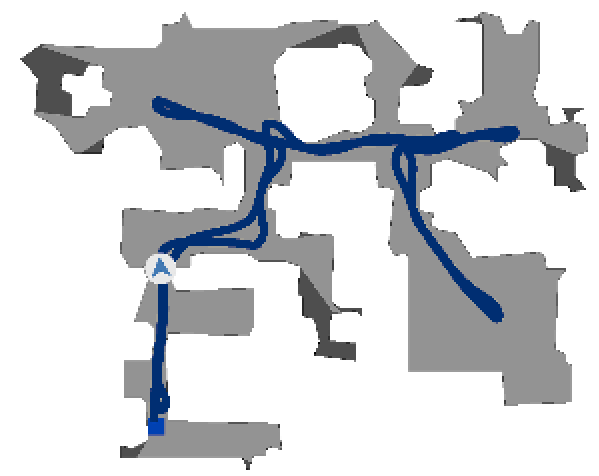}
        \caption{}
    \end{subfigure}
    \hfill
    \begin{subfigure}{0.4\linewidth}
        \centering
        \includegraphics[width=\linewidth]{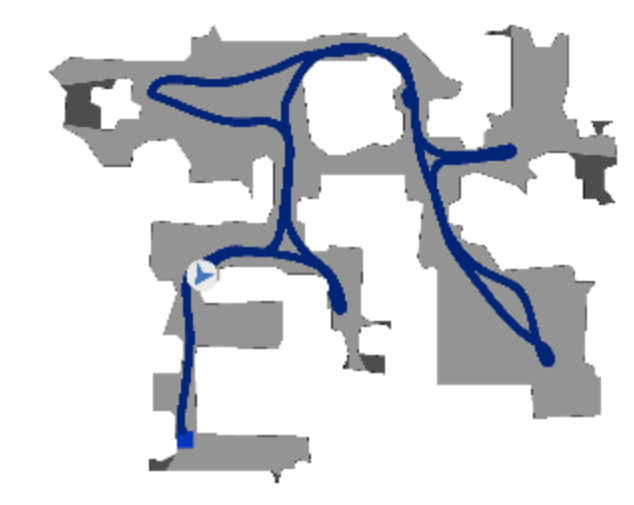}
        \caption{}
    \end{subfigure}
    \caption{The robot's exploration paths for Environment 1 for all five exploration methods. (a) Ours, (b) our method with a scale shown in red, and green points indicating locations where it performed a 360 \textdegree{} scan, (c) Greedy, (d) OpenCV + NBV, (e) TARE, and (f) DSVP. Light gray represents explored areas, dark gray represents unexplored areas, white represents out-of-bounds or untraversable area, and the exploration paths are shown in blue.}
    \label{fig:final_pics_env_1}
\end{figure}

\begin{figure}[htbp!]
    \begin{subfigure}{0.49\linewidth}
        \centering
        \includegraphics[width=\linewidth]{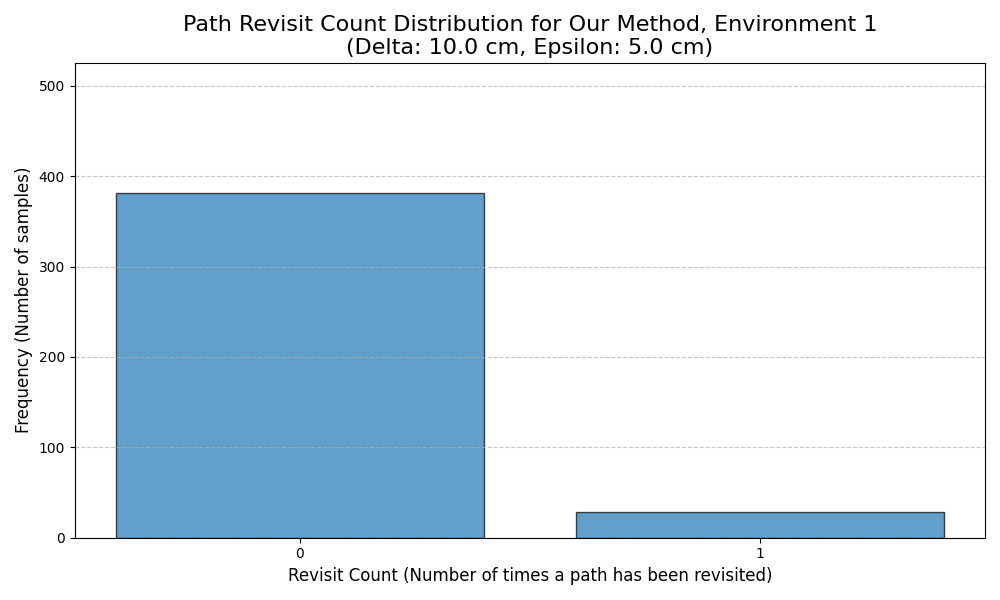}
        \caption{}
    \end{subfigure}
    \begin{subfigure}{0.49\linewidth}
        \centering
        \includegraphics[width=\linewidth]{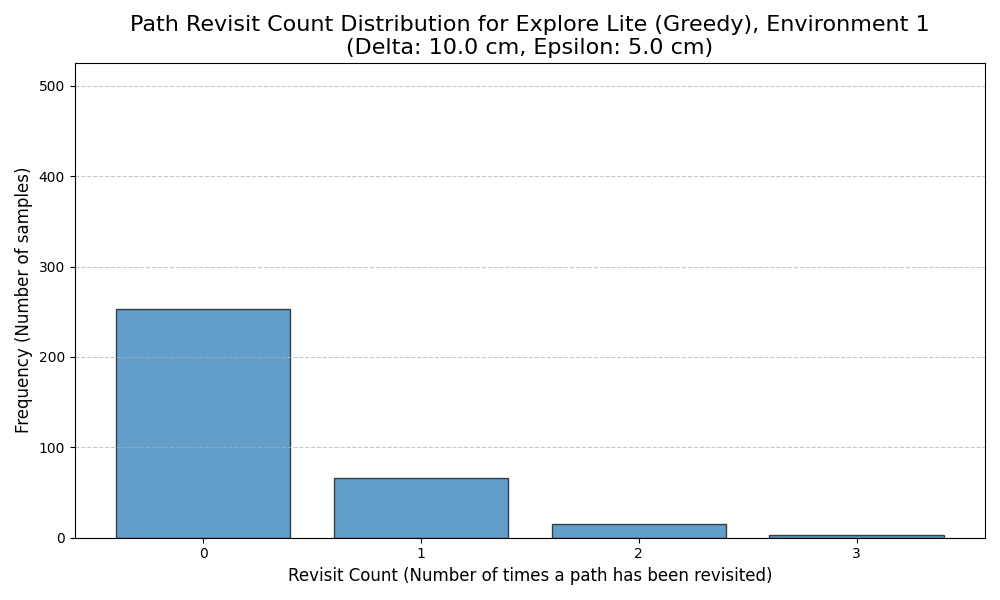}
        \caption{}
    \end{subfigure}
    \begin{subfigure}{0.49\linewidth}
        \centering
        \includegraphics[width=\linewidth]{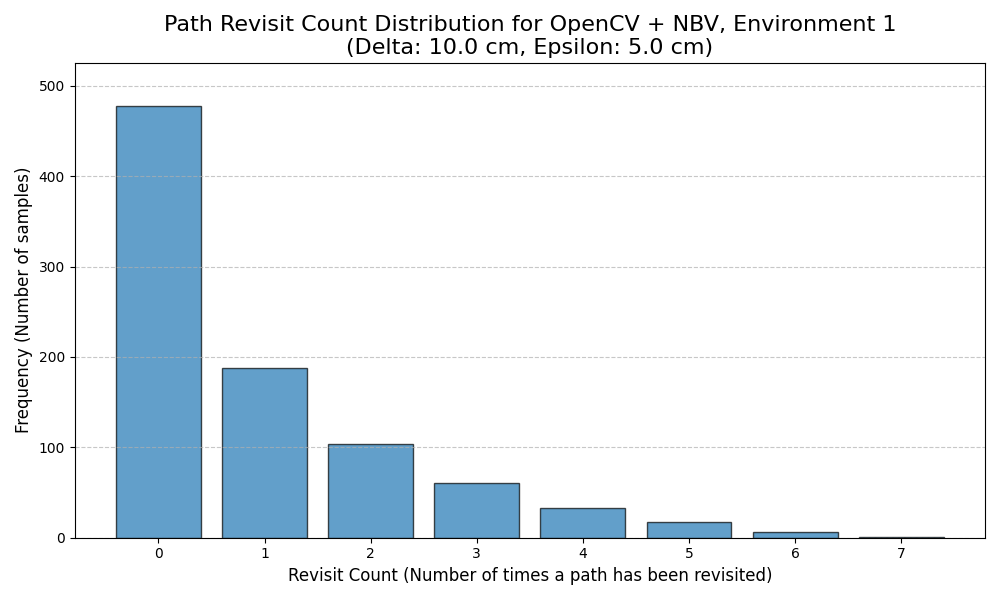}
        \caption{}
    \end{subfigure}
    \begin{subfigure}{0.49\linewidth}
        \centering
        \includegraphics[width=\linewidth]{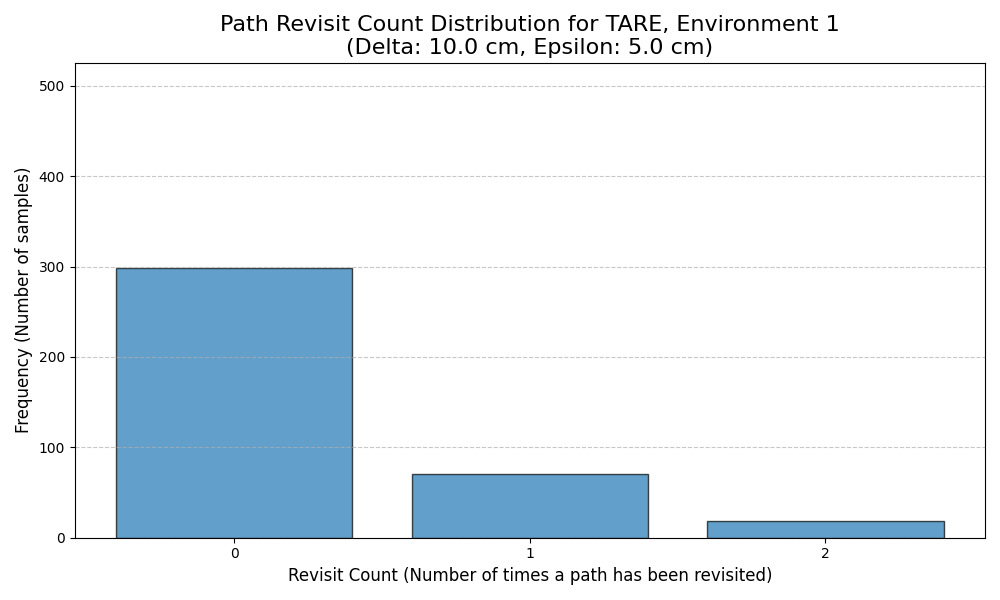}
        \caption{}
    \end{subfigure}
    \begin{subfigure}{0.49\linewidth}
        \centering
        \includegraphics[width=\linewidth]{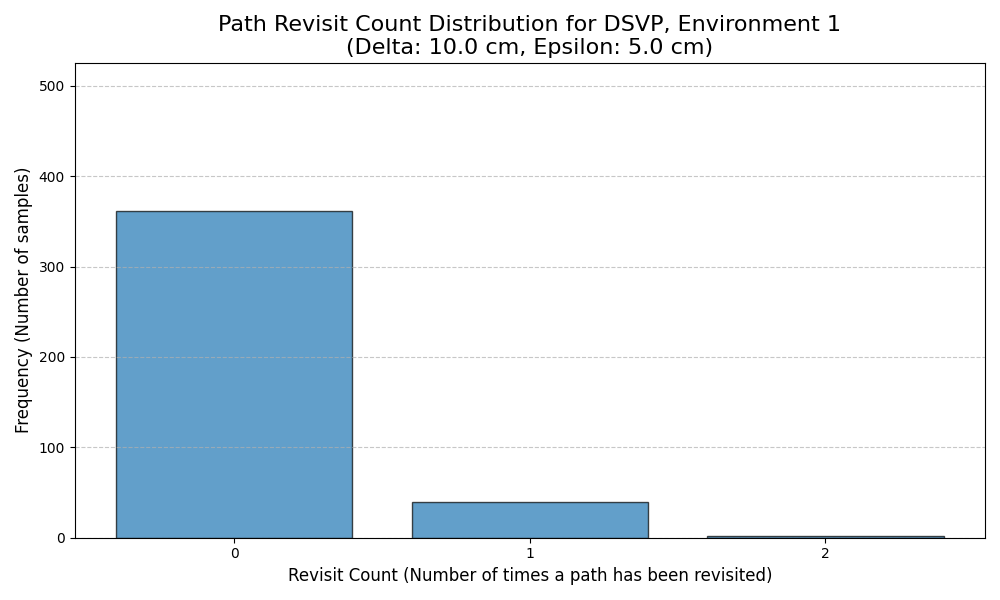}
        \caption{}
    \end{subfigure}
    \caption{Histograms of path revisit counts for Environment 1. (a) Ours, (b) Greedy, (c) OpenCV + NBV, (d) TARE, and (e) DSVP.}
    \label{fig:trajectory_repetition_env_1}
\end{figure}

\begin{figure}[htbp!]
    \begin{subfigure}{0.35\linewidth}
        \centering
        \includegraphics[width=\linewidth]{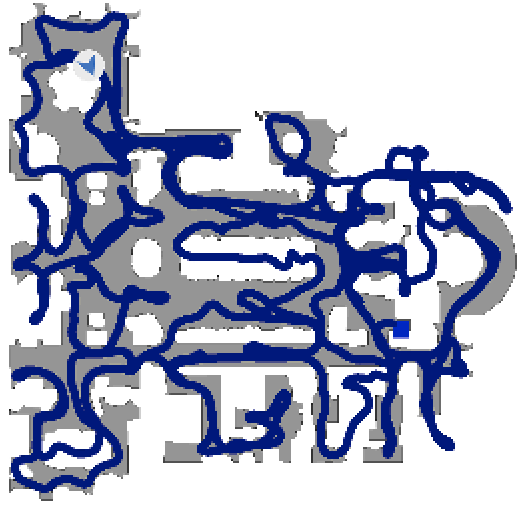}
        \caption{}
    \end{subfigure}
    \hfill
    \begin{subfigure}{0.35\linewidth}
        \centering
        \includegraphics[width=\linewidth]{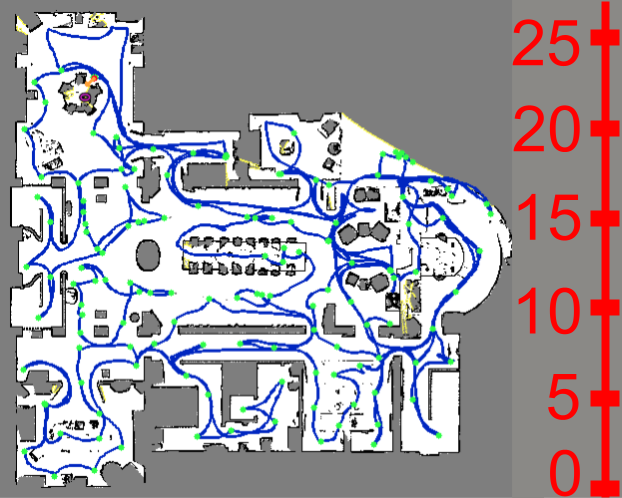}
        \caption{}
    \end{subfigure}
    \begin{subfigure}{0.35\linewidth}
        \centering
        \includegraphics[width=\linewidth]{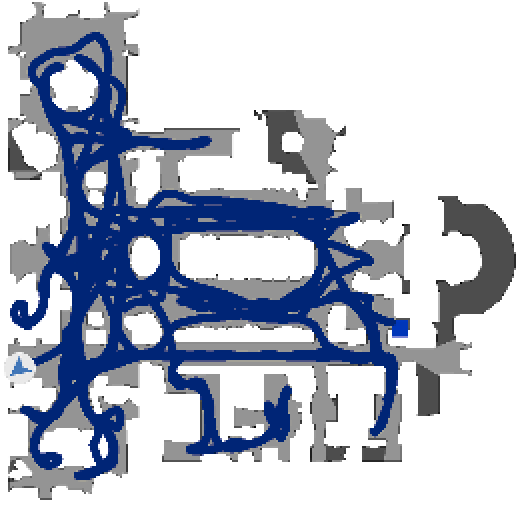}
        \caption{}
    \end{subfigure}
    \begin{subfigure}{0.35\linewidth}
        \centering
        \includegraphics[width=\linewidth]{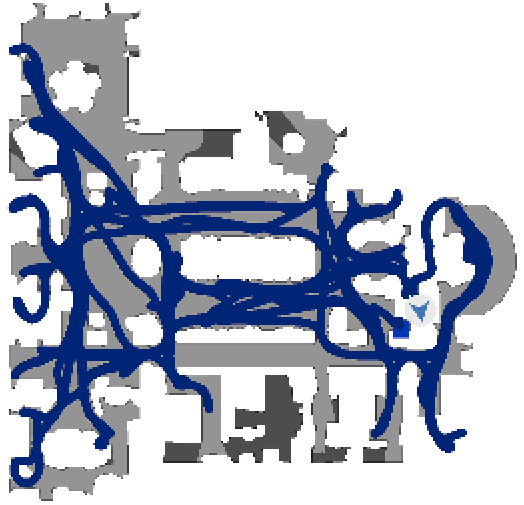}
        \caption{}
    \end{subfigure}
    \begin{subfigure}{0.35\linewidth}
        \centering
        \includegraphics[width=\linewidth]{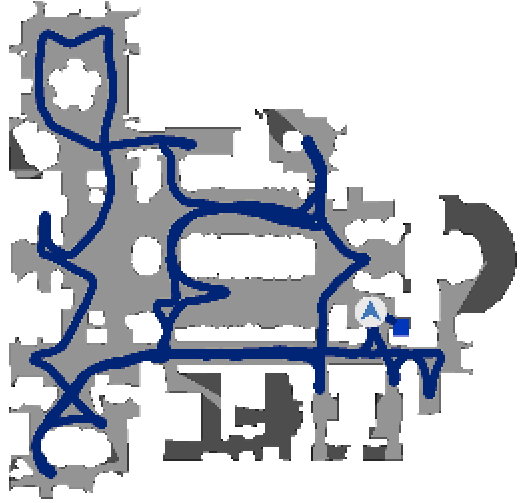}
        \caption{}
    \end{subfigure}
    \hfill
    \begin{subfigure}{0.35\linewidth}
        \centering
        \includegraphics[width=\linewidth]{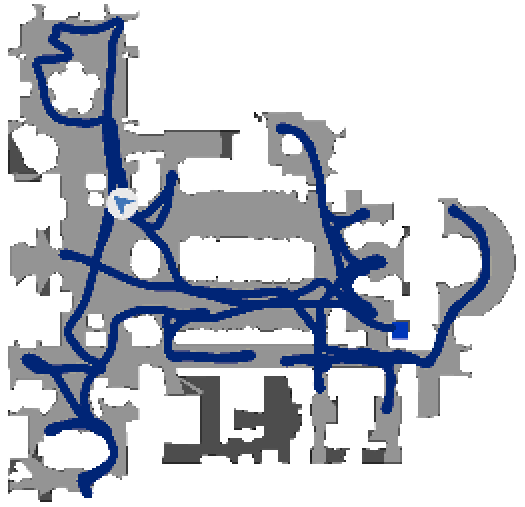}
        \caption{}
    \end{subfigure}
    \caption{The robot's exploration paths for Environment 2 for all five exploration methods. (a) Ours, (b) our method with a scale in meters shown in red, and green points indicating locations where it performed a 360 \textdegree{} scan, (c) Greedy, (d) OpenCV + NBV, (e) TARE, and (f) DSVP. Light gray represents explored areas, dark gray represents unexplored areas, white represents out-of-bounds or untraversable area, and the exploration paths are shown in blue.}
    \label{fig:final_pics_env_2}
\end{figure}

\begin{figure}[htbp!]
    \begin{subfigure}{0.49\linewidth}
        \centering
        \includegraphics[width=\linewidth]{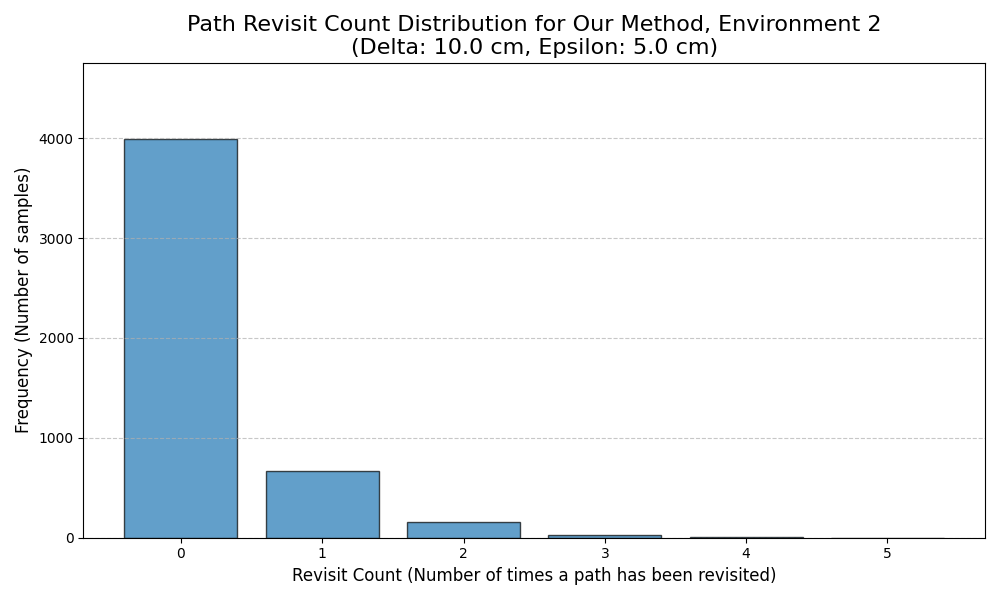}
        \caption{}
    \end{subfigure}
    \begin{subfigure}{0.49\linewidth}
        \centering
        \includegraphics[width=\linewidth]{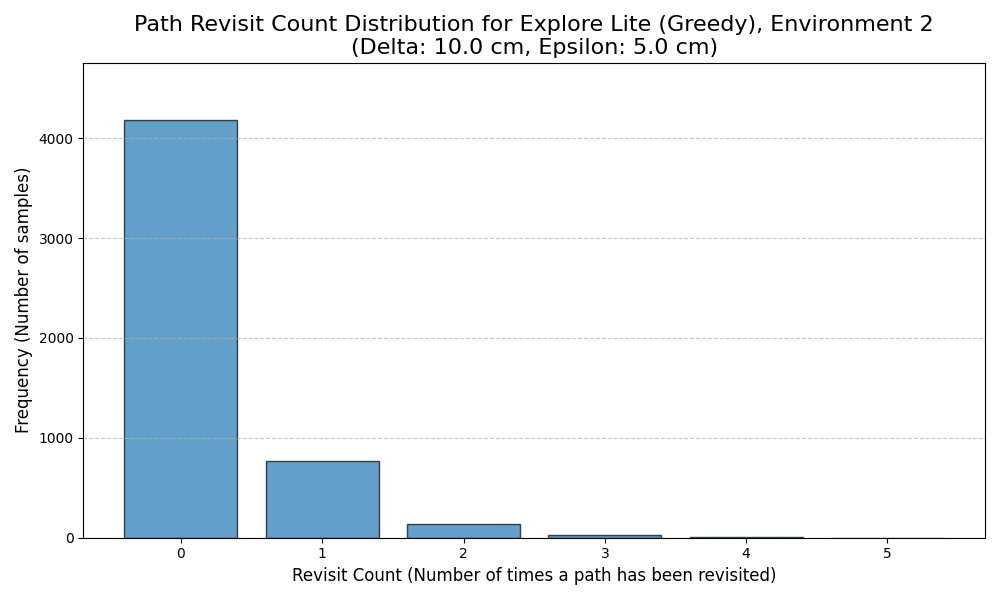}
        \caption{}
    \end{subfigure}
    \begin{subfigure}{0.49\linewidth}
        \centering
        \includegraphics[width=\linewidth]{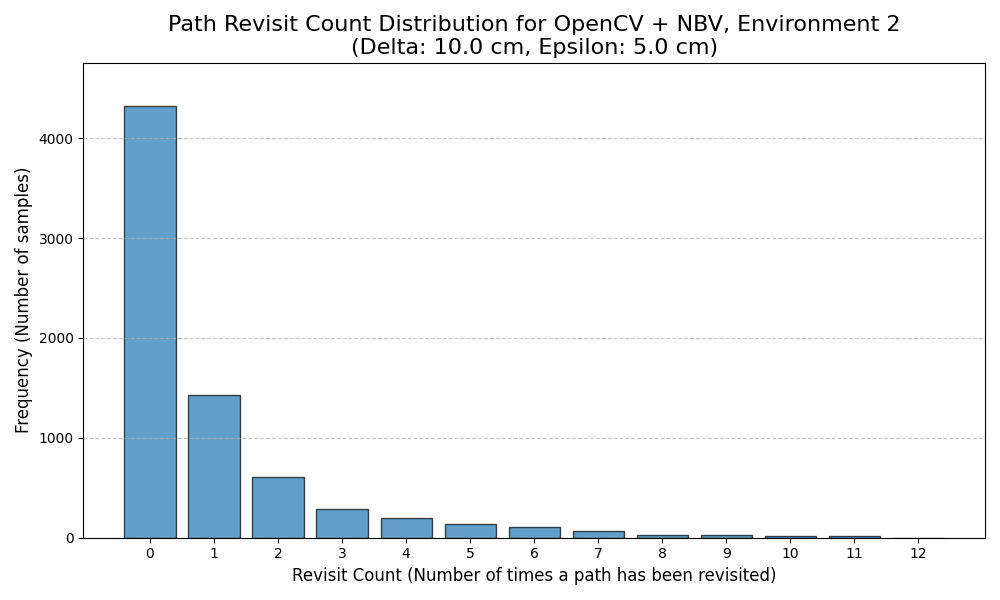}
        \caption{}
    \end{subfigure}
    \begin{subfigure}{0.49\linewidth}
        \centering
        \includegraphics[width=\linewidth]{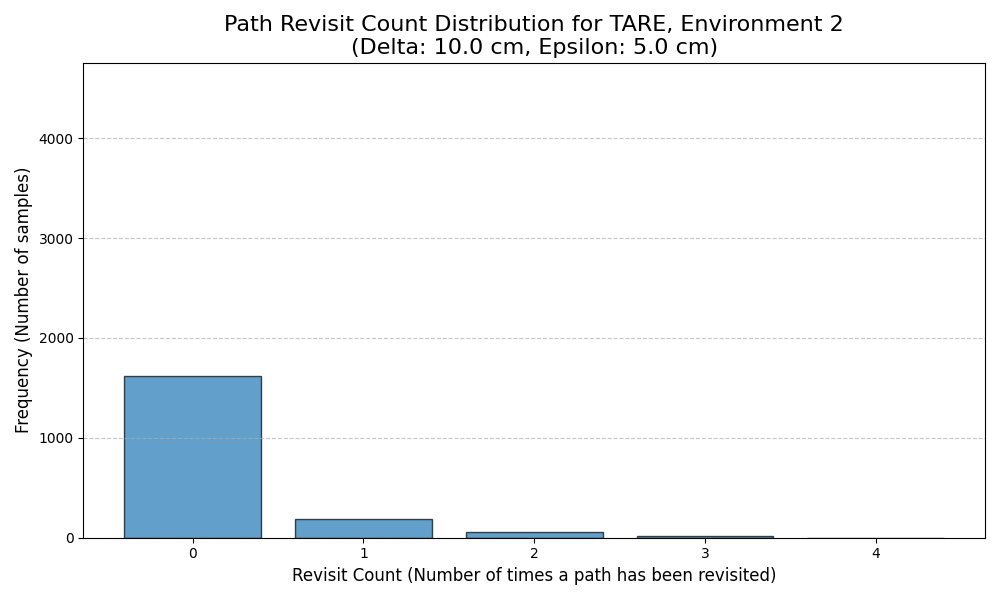}
        \caption{}
    \end{subfigure}
    \begin{subfigure}{0.49\linewidth}
        \centering
        \includegraphics[width=\linewidth]{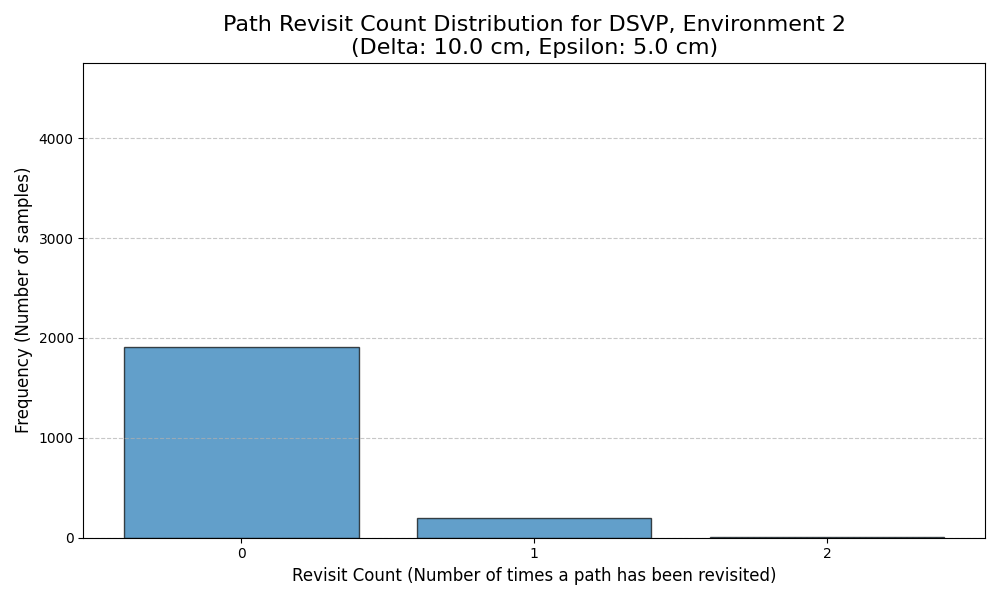}
        \caption{}
    \end{subfigure}
    \caption{Histograms of path revisit counts for Environment 2. (a) Ours, (b) Greedy, (c) OpenCV + NBV, (d) TARE, and (e) DSVP.}
    \label{fig:trajectory_repetition_env_2}
\end{figure}

The quantitative data is supported by a qualitative review of the final exploration maps and trajectories. For the sake of brevity, we show the qualitative results for environments 1 and 2 only, as shown in Figures \ref{fig:final_pics_env_1} and \ref{fig:final_pics_env_2} respectively. We additionally show histograms of the repeated traversals in the exploration paths, shown in Figures \ref{fig:trajectory_repetition_env_1} and \ref{fig:trajectory_repetition_env_2} for environments 1 and 2 respectively. To create these histograms, we first process the 2D trajectory of each method by resampling the path into a sequence of points with a uniform spacing of 10 cm. Then, we iterate through each point in this new sequence to calculate its 'revisit count'. For a given point, we search for any points that appeared earlier in the trajectory and fall within a 5 cm radius, excluding the immediate preceding points of the current path segment. The revisit count for the current point is then defined as one plus the maximum revisit count found among these past neighboring points. If no past points are found within the radius, the point is considered part of a new path segment and is assigned a revisit count of zero. The resulting histograms show the distribution of these counts, with the x-axis representing the number of times a path segment has been revisited and the y-axis representing the frequency of points with that count.
\\
\\
In Figure \ref{fig:final_pics_env_1}, all methods successfully map the relatively simple environment, but the path taken by our method, shown in \ref{fig:final_pics_env_1}(a) is visibly more direct than the meandering paths of the Greedy \ref{fig:final_pics_env_1}(c) and NBV \ref{fig:final_pics_env_1}(d) planners. We can see that the histograms for those two methods in Figures \ref{fig:trajectory_repetition_env_1}(b) and \ref{fig:trajectory_repetition_env_1}(c) both have many more points with higher path repetitions compared to our method, shown in \ref{fig:trajectory_repetition_env_1}(a).

In the more complex Environment 2, shown in Figure \ref{fig:final_pics_env_2}, the weaknesses of the baselines are evident. The Greedy and NBV planners fail to fully explore the large circular room on the right, while TARE and DSVP terminate without ever visiting the top portion of the map. Additionally, the Greedy and NBV planners both repeatedly traverse long hallways, causing their total distance traveled to be much longer than necessary. Our method is the only one to achieve comprehensive coverage while also minimizing backtracking distance. The histograms in Figures \ref{fig:trajectory_repetition_env_2}(b) and \ref{fig:trajectory_repetition_env_2}(c) support these claims, as they have many points in bins 1 and higher, i.e. the paths were re-traversed many times. This is juxtaposed with our method's histogram in Figure \ref{fig:trajectory_repetition_env_2}(a), which has the majority of its points in unique paths while a small fraction are repeated points due to having to backtrack out of dead-ends.

Although not shown, a consistent trend was observed across the remaining environments \cite{aitha2025frontier}.

\begin{itemize}
\item Environment 3: In this smaller space, most methods achieved high coverage. However, the TARE and DSVP planners tended to over-explore, retraversing paths multiple times before termination, unlike our method which stopped efficiently once mapping was complete \cite{aitha2025frontier}.
\item Environments 4 \& 5: The premature termination of TARE and DSVP was particularly evident in these medium-to-large environments, where they left entire wings of the map unexplored. Our VLM-guided agent, by contrast, successfully navigated these larger spaces to achieve near-full coverage \cite{aitha2025frontier}.
\item Environment 6: Our method again demonstrated a smooth and efficient trajectory. In contrast, the Greedy and TARE planners were observed repeating loops in the central area of the map, while the NBV planner failed to explore a significant portion of the environment entirely \cite{aitha2025frontier}.
\end{itemize}
Across all test cases, the VLM-guided agent's paths were more strategically sound and comprehensive, avoiding the common pitfalls of premature termination and inefficient path repetition shown in other methods.

\section{CONCLUSIONS}

This work demonstrated that integrating a VLM into a conventional robotics stack significantly enhances autonomous exploration. By delegating strategic planning to the VLM, our framework supplants rigid geometric heuristics with nuanced spatial reasoning, allowing the robot to select frontiers based on a holistic interpretation of the environment's topology. Our VLM-driven agent consistently outperformed established methods, including greedy, NBV, and hierarchical planners like TARE and DSVP. This research validates the hypothesis that the emergent spatial reasoning of VLMs can address complex robotics tasks, outlining a paradigm for more adaptable and context-aware autonomous agents. We additionally attempted a preliminary sim-to-real transfer of our pipeline onto a physical robot; for details on this effort, the reader is referred to the full thesis \cite{aitha2025frontier}.







\end{document}